\documentstyle[ling,cgloss4e,qtree,aclap]{article}

\qtreecenterfalse                            % don't center trees in available space

\newtheorem{theorem}{Theorem}

\newcommand{\refp}[1]{(\ref{#1})}   % this one from Alexis.sty
\newcommand{\litem}[1]{\item\label{#1}}   % from Alexis.sty, but I removed the margin note
\newcommand{\longcomment}[1]{}

\newcommand{\bs}{\mbox{$\backslash$}}
\newcommand{\slsh}[1]{\mbox{$\,\mid_#1$}}   % subscripted vertical bar with no space around it
\newcommand{\cat}[1]{{\tt {#1}}}
\newcommand{\subcat}[1]{\raisebox{-0.5ex}{\scriptsize #1}}
\newcommand{\tag}[1]{\mbox{\sc --#1}}
\newcommand{\cattag}[2]{\ \ \ \cat{#1}\makebox[0in][l]{\tag{#2}}\ \ \ }
\newcommand{\catrule}[2]{\ \ \ \ \ \ \ \cat{#1}\makebox[0in][l]{\small\ (\cat{#2})}\ \ \ \ \ \ \ }

\newcommand{\catword}[2]{$\begin{array}[t]{c}\mbox{\cat{#1}} \\ \mbox{\small #2}\end{array}$}
\newcommand{\Trees}[1]{\mbox{\def\baselinestretch{1}\small\normalsize\Tree #1 \hspace{-0.5in}}}  % my wrapper on \Tree, which enforces single spacing etc -- can embed this tree in running text -- also, it kills nasty post-tree space.  
\newcommand{\ang}[1]{{\mbox{$<$}#1\mbox{$>$}}}   % surrounds with nice angle brackets 
\newcommand{\chooz}[2]{\mbox{\scriptsize $\left(\begin{array}{@{}c@{}}{#1} \\ {#2} \end{array} \right)$ \normalsize}}   % choose operator
\newcommand{\alternatives}[2]{\( \left\{ \!\! \begin{array}{#1} #2 \end{array} \!\! \right\} \)}
\newcommand{\qed}{\hfill\rule{0.6em}{0.6em}}
\newcommand{\tb}{\hspace{0.16in}}

\newcommand{\treeexample}[1]{\hspace{-#1\leftmargin}}

\title{\vspace{-0.5in}Efficient Normal-Form Parsing \\ for Combinatory Categorial Grammar%
\thanks{This material is based upon work supported under a National
Science Foundation Graduate Fellowship.  I have been grateful for the advice of Aravind 
Joshi, Nobo Komagata, Seth Kulick, Michael Niv, Mark Steedman, and three
anonymous reviewers.}}
\author{Jason Eisner \\
Dept.\ of Computer and Information Science \\
University of Pennsylvania \\
200 S. 33rd St., Philadelphia, PA  19104-6389, USA \\{\tt jeisner@linc.cis.upenn.edu}}

\begin{document}
\bibliographystyle{fullname}
\maketitle
\vspace{-0.5in}
\begin{abstract}

Under categorial grammars that have powerful rules like composition, a
simple $n$-word sentence can have exponentially many parses.
Generating all parses is inefficient and obscures whatever true
semantic ambiguities are in the input.  This paper addresses the
problem for a fairly general form of Combinatory Categorial Grammar,
by means of an efficient, correct, and easy to implement
normal-form parsing technique.  The parser is proved to find {\em
exactly one} parse in each semantic equivalence class of allowable
parses; that is, spurious ambiguity (as carefully defined) is shown to
be both safely and completely eliminated.
\end{abstract}

\section{Introduction}

Combinatory Categorial Grammar \cite{Steedman90}, like other
``flexible'' categorial grammars, \linebreak suffers from {\em
spurious ambiguity} \cite{Wittenburg86}.  The non-standard
constituents that are so crucial to CCG's analyses in
\refp{nonstandard}, and in its account of intonational focus
\cite{Prevost+Steedman}, remain available even in simpler sentences.
This renders \refp{spuriousambig} syntactically ambiguous.

\begin{raggedright}
\bsent
   \litem{nonstandard}\subsent
	\litem{nonstandard:coord} {\bf Coordination:} [[\underline{John likes}]\subcat{S/NP}, and [Mary
pretends to like]\subcat{S/NP$\:$}], the big \mbox{galoot} in the corner.
	\litem{nonstandard:extract} {\bf Extraction:} Everybody at this party [whom [\underline{John likes}]\subcat{S/NP$\:$}]
is a big galoot.
    \esubsent
\medskip
    \litem{spuriousambig}\subsent
        \litem{spuriousambig:standard} John [likes Mary]\subcat{S\bs NP}.
        \litem{spuriousambig:nonstandard} [\underline{John likes}]\subcat{S/NP} Mary.
   \esubsent
\esent
\end{raggedright}

The practical problem of ``extra'' parses in \refp{spuriousambig} becomes exponentially worse for
longer strings, which can have up to a Catalan number of parses.  An
exhaustive parser serves up 252 CCG parses of \refp{lotsofspurious},
which must be sifted through, at considerable cost, in order to
identify the two distinct meanings for further processing.%
\footnote{Namely, Mary pretends to like the galoot in 168 parses and
the corner in 84.  One might try a statistical approach to ambiguity
resolution, discarding the low-probability parses, but it is unclear
how to model and train any probabilities when no single parse can be
taken as the standard of correctness.}

\begin{raggedright}
\bsent 
   \litem{lotsofspurious} \gll the  galoot in          the  corner that         I  said  Mary     pretends               to                            like \\
                               {\tiny NP/N\ \ } {\tiny N\ \ }     {\tiny (N\bs N)/NP\ \ } {\tiny NP/N\ \ } {\tiny N\ \ }   {\tiny (N\bs N)/(S/NP)\ \ } {\tiny S/(S\bs NP)\ \ } {\tiny (S\bs NP)/S\ \ } {\tiny S/(S\bs NP)\ \ } {\tiny (S\bs NP)/(S$_{\rm inf}$\bs NP)\ \ } {\tiny (S$_{\rm inf}$\bs NP)/(S$_{\rm stem}$\bs NP)\ \ } {\tiny (S$_{\rm stem}$\bs NP)/NP\ \ } \\  
\esent
\end{raggedright}

This paper presents a simple and flexible CCG parsing technique that
prevents any such explosion of redundant CCG derivations.  In
particular, it is proved in \S\ref{sec:proofs} that the method
constructs {\em exactly one} syntactic structure per semantic
reading---e.g., just two parses for \refp{lotsofspurious}.  All other
parses are suppressed by simple normal-form constraints that are
enforced throughout the parsing process.  This approach works because
CCG's spurious ambiguities arise (as is shown) in only a small set of 
circumstances.  Although similar work has been attempted in the past,
with varying degrees of success
\cite{Karttunen86,Wittenburg86,Pareschi+Steedman,Bouma89,Hepple+Morrill89,%
Konig89,VijayWeirPolyparseSpur,Hepple90thesis,Moortgat90,Hendriks93,Niv94},
this appears to be the first full normal-form result for a categorial
formalism having more than context-free power.

\section{Definitions and Related Work}\label{ccgdefs}

CCG may be regarded as a generalization of context-free grammar
(CFG)---one where a grammar has infinitely many nonterminals and
phrase-structure rules.  In addition to the familiar {\em atomic}
nonterminal categories (typically \cat{S} for sentences, \cat{N} for
nouns, \cat{NP} for noun phrases, etc.), CCG allows infinitely many
{\em slashed} categories.  If $x$ and $y$ are categories, then $x/y$
(respectively $x\bs y$) is the category of an incomplete $x$ that is
missing a $y$ at its right (respectively left).  Thus verb phrases are
analyzed as subjectless sentences \cat{S\bs NP}, while ``John likes''
is an objectless sentence or \cat{S/NP}.  A complex category like
\cat{((S\bs NP)\bs(S\bs NP))/N} may be written as \cat{S\bs NP\bs(S\bs
NP)/N}, under a convention that slashes are left-associative.

The results herein apply to the TAG-equivalent CCG formalization given
in \cite{Joshi+al}.%
\footnote{This formalization sweeps any type-raising into the lexicon,
as has been proposed on linguistic grounds \cite[and
others]{DowtyLexTR,SteedmanLexTR}.  It also treats conjunction
lexically, by giving ``and'' the generalized category $x\bs x/x$ and
barring it from composition.}  In this variety of CCG, every
(non-lexical) phrase-structure rule is an instance of one of the
following binary-rule templates (where $n \geq 0$):

\begin{raggedright}
\bsent
   \litem{template}  % we'll use negative hspace to de-indent the rules so they fit
          Forward generalized composition \cat{>B}$n$:
          \\ \hspace{-\leftmargin}\mbox{}\hfill $x/y\hfill y\slsh{n}z_n\cdots\slsh{2}z_2\slsh{1}z_1   
          \ \rightarrow\ x\slsh{n}z_n\cdots\slsh{2}z_2\slsh{1}z_1$
       \\ Backward generalized composition \cat{<B}$n$:
          \\ \hspace{-\leftmargin}\mbox{}\hfill $y\slsh{n}z_n\cdots\slsh{2}z_2\slsh{1}z_1\hfill x\bs y
          \ \rightarrow \ x\slsh{n}z_n\cdots\slsh{2}z_2\slsh{1}z_1$
\esent
\end{raggedright}
Instances with $n=0$ are called {\em application} rules, and instances with $n\geq 1$ are called {\em composition} rules.  In a given rule, 
$x, y, z_1 \ldots z_n$ would be instantiated as categories like
\cat{NP}, \cat{S/NP}, or \cat{S\bs NP\bs(S\bs NP)/N}.
Each of $\slsh{1}$~through~$\slsh{n}$ would be instantiated as
either \cat{/} or \cat{\bs}.  

A fixed CCG grammar need not include every phrase-structure rule
matching these templates.  Indeed, \cite{Joshi+al} place certain
restrictions on the rule set of a CCG grammar, including a requirement
that the rule degree $n$ is bounded over the set.  The results of the
present paper apply to such restricted grammars and also more
generally, to any CCG-style grammar with a {\em decidable} rule set.

Even as restricted by \cite{Joshi+al}, CCGs have the ``mildly
context-sensitive'' expressive power of Tree Adjoining Grammars
(TAGs).  Most work on spurious ambiguity has focused on categorial
formalisms with substantially less power.  \cite{Hepple90thesis} and
\cite{Hendriks93}, the most rigorous pieces of work, each establish a
normal form for the syntactic calculus of \cite{Lambek}, which is
weakly context-free.  \cite{Konig89,Moortgat90} have also studied the
Lambek calculus case.  \cite{Hepple+Morrill89}, who introduced the
idea of normal-form parsing, consider only a small CCG fragment that
lacks backward or order-changing composition; \cite{Niv94} extends
this result but does not show completeness.  \cite{Wittenburg87}
assumes a CCG fragment lacking order-changing or higher-order
composition; furthermore, his revision of the combinators creates new,
conjoinable
constituents that conventional CCG rejects.  \cite{Bouma89} proposes
to replace composition with a new combinator, but the resulting
product-grammar scheme assigns different types to ``John likes'' and
``Mary pretends to like,'' thus losing the ability to conjoin such
constituents or subcategorize for them as a class.
\cite{Pareschi+Steedman} do tackle the CCG case, but \cite{Hepple87}
shows their algorithm to be incomplete.  

\section{Overview of the Parsing Strategy}\label{overview}

As is well known, general CFG parsing methods can be applied directly
to CCG.  Any sort of chart parser or non-deterministic shift-reduce
parser will do.  Such a parser repeatedly decides whether two adjacent
constituents, such as \cat{S/NP} and \cat{NP/N}, should be combined
into a larger constituent such as \cat{S/N}.  The role of the grammar
is to state which combinations are allowed.  The key to efficiency, we
will see, is for the parser to be less permissive than the
grammar---for it to say ``no, redundant'' in some cases where the 
grammar says ``yes, grammatical.''

\refp{constits} shows the constituents that untrammeled CCG will find
in the course of parsing ``John likes Mary.''  The spurious ambiguity
problem is not that the grammar allows \refp{constits:Johnlikes}, but
that the grammar allows both \refp{constits:nonnf} and
\refp{constits:nf}---distinct parses of the {\em same} string, with
the {\em same} meaning.

\begin{raggedright}
\bsent
   \litem{constits}\subsent
      \litem{constits:John} [John]\subcat{S/(S\bs NP)}
      \litem{constits:likes} [likes]\subcat{(S\bs NP)/NP}
      \litem{constits:Johnlikes} [John likes]\subcat{S/NP}
      \litem{constits:Mary} [Mary]\subcat{NP}
      \litem{constits:likesMary} [likes Mary]\subcat{S\bs NP}
      \litem{constits:nonnf} [[John likes] Mary]\subcat{S} \hfill {\small \em $\longleftarrow$ to be disallowed}
      \litem{constits:nf} [John [likes Mary]]\subcat{S}
   \esubsent
\esent
\end{raggedright}

The proposal is to construct all constituents shown in \refp{constits}
except for \refp{constits:nonnf}.  If we slightly constrain the use of
the grammar rules, the parser will still produce
\refp{constits:Johnlikes} and \refp{constits:Mary}---constituents that
are indispensable in contexts like \refp{nonstandard}---while refusing
to {\em combine} those constituents into \refp{constits:nonnf}.  The
relevant rule \cat{S/NP} \cat{NP} $\rightarrow$ \cat{S} will actually
be blocked when it attempts to construct \refp{constits:nonnf}.
Although rule-blocking may eliminate an analysis of the sentence, as
it does here, a semantically equivalent analysis such as
\refp{constits:nf} will always be derivable along some other route.

In general, our goal is to discover exactly one analysis for each
\ang{substring, meaning} pair.  By practicing ``birth control'' for
each bottom-up generation of constituents in this way, we avoid a
population explosion of parsing options.  ``John likes Mary'' has only
one reading semantically, so just one of its analyses
\refp{constits:nonnf}--\refp{constits:nf} is discovered while parsing
\refp{noexplode}.  Only that analysis, and not the other, is allowed
to continue on and be built into the final parse of \refp{noexplode}.

\begin{raggedright}
\bsent
   \litem{noexplode} that galoot in the corner that thinks [John likes Mary]\subcat{S}
\esent
\end{raggedright}

For a chart parser, where each chart cell stores the analyses of some
substring, this strategy says that all analyses in a cell are to be
semantically distinct.  \cite{Karttunen86} suggests enforcing that
property directly---by comparing each new analysis semantically with
existing analyses in the cell, and refusing to add it if redundant---but
\cite{Hepple+Morrill89} observe briefly that this is inefficient for
large charts.%
\footnote{\label{note:karttunen}How inefficient?
(i) has exponentially many semantically {\em distinct}
parses: $n=10$ yields 82,756,612 parses in \chooz{20}{10} = 48,620
equivalence classes.  Karttunen's method must therefore add 48,620
representative parses to the appropriate chart cell, first comparing
each one against all the previously added parses---of which there
are 48,620/2 on average---to ensure it is not semantically redundant.  
(Additional comparisons are needed to reject parses other than the lucky 48,620.)
Adding a parse can therefore take exponential time.

\begin{raggedright}
\bsent
   \item[(i)] $\overbrace{\ldots\ \cat{S/S}\ \ \cat{S/S}\ \ \cat{S/S}}^n\ \ \cat{S}\ \ \overbrace{\cat{S\bs S}\ \ \cat{S\bs S}\ \ \cat{S\bs S}\ \ldots}^n$
\esent
\end{raggedright}
Structure sharing does not appear to help: parses that are grouped in
a parse forest have only their syntactic category in common, not their
meaning.  Karttunen's approach must tease such parses apart and
compare their various meanings individually against each new
candidate.  By contrast, the method proposed below is purely
syntactic---just like any ``ordinary'' parser---so it never needs to
unpack a subforest, and can run in polynomial time.}
The following sections show how to obtain effectively the same
result without doing any semantic interpretation or comparison at
all.

\section{A Normal Form for ``Pure'' CCG}\label{sec:nfparser}

It is convenient to begin with a special case.  Suppose the CCG
grammar includes not some but {\em all} instances of the binary rule
templates in \refp{template}.  (As always, a separate lexicon
specifies the possible categories of each word.)  If we group a
sentence's parses into semantic equivalence classes, it always turns
out that exactly one parse in each class satisfies the following
simple declarative constraints:

\begin{raggedright}
\bsent
   \litem{constraints}\subsent
      \litem{constraints:forward} No constituent produced by \cat{>B}$n$, any $n \geq 1$, ever serves
as the primary (left) argument to \cat{>B}$n'$, any $n' \geq 0$.
      \litem{constraints:backward} No constituent produced by \cat{<B}$n$, any $n \geq 1$, ever serves
as the primary (right) argument to \cat{<B}$n'$, any $n' \geq 0$.
   \esubsent
\esent
\end{raggedright}

The notation here is from \refp{template}.  More colloquially,
\refp{constraints} says that the output of rightward (leftward)
composition may not compose or apply over anything to {\em its} right
(left).  A parse tree or subtree that satisfies \refp{constraints} is
said to be in {\em normal form} (NF).

As an example, consider the effect of these restrictions on the simple
sentence ``John likes Mary.''  Ignoring the tags \tag{ot}, \tag{fc},
and \tag{bc} for the moment, \refp{nfexamples:nf} is a normal-form parse.
Its competitor \refp{nfexamples:nonnf} is not, nor is any larger tree
containing \refp{nfexamples:nonnf}.  But non-standard constituents are
allowed when necessary: \refp{nfexamples:nfnonstandard} {\em is} in
normal form (cf. \refp{nonstandard}).

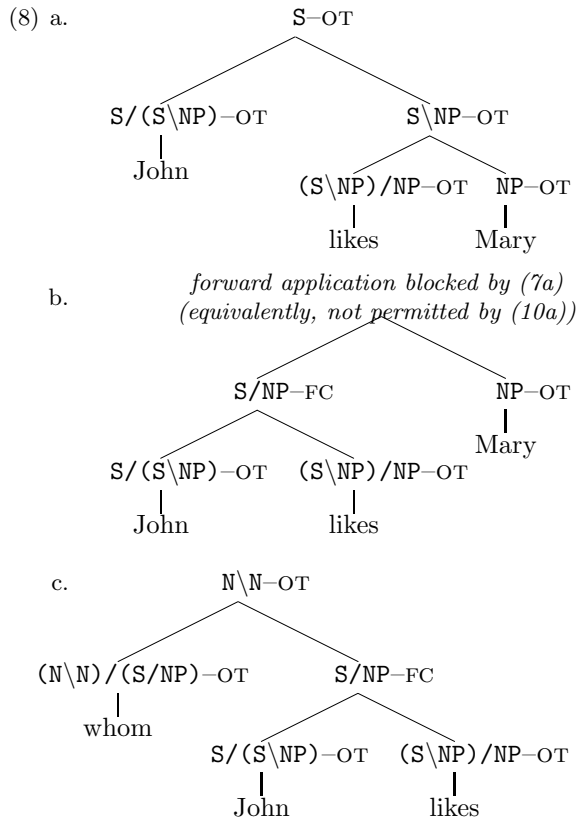
\begin{raggedright}
\bsent \small
   \litem{nfexamples}\subsent
      \litem{nfexamples:nf} \treeexample{2}%
\Trees{[ [ John ].\cattag{S/(S\bs NP)}{ot} [ [ likes ].\cattag{(S\bs NP)/NP}{ot}  [ Mary ].\cattag{NP}{ot} ].\cattag{S\bs NP}{ot} ].\cattag{S}{ot}}
\bigskip
      \litem{nfexamples:nonnf} \treeexample{2}%
\mbox{\Trees{[ [ [ John ].\cattag{S/(S\bs NP)}{ot} [ likes ].\cattag{(S\bs NP)/NP}{ot} ].\cattag{S/NP}{fc} [ Mary ].\cattag{NP}{ot} ].\raisebox{0in}[0in][0in]{\parbox{6cm}{\center {\small \em forward application blocked by \refp{constraints:forward}}\ \ \ \\ {\small \em (equivalently, not permitted by \refp{constrainedtemplate:fa})}\ \ \ \\}}} }
\bigskip
      \litem{nfexamples:nfnonstandard} \treeexample{2}%
\Trees{[ [ whom ].\makebox[0in][c]{\cattag{(N\bs N)/(S/NP)}{ot}} [ [ John ].\cattag{S/(S\bs NP)}{ot} [ likes ].\cattag{(S\bs NP)/NP}{ot} ].\cattag{S/NP}{fc} ].\cattag{N\bs N}{ot}}
   \esubsent
\esent
\end{raggedright}

It is not hard to see that \refp{constraints:forward} eliminates all
but right-branching parses of ``forward chains'' like \cat{A/B $\,$B/C $\,$C}
or \cat{A/B/C $\,$C/D $\,$D/E/F/G $\,$G/H}, and that \refp{constraints:backward}
eliminates all but left-branching parses of ``backward chains.''
(Thus every functor will get its arguments, if possible, before it
becomes an argument itself.)  But it is hardly obvious that
\refp{constraints} eliminates {\em all} of CCG's spurious ambiguity.
One might worry about unexpected interactions involving crossing
composition rules like \mbox{\cat{A/B $\,$B\bs C} $\rightarrow$ \cat{A\bs
C}}.  Significantly, it turns out that \refp{constraints} really does
suffice; the proof is in \S\ref{sec:proofs}.

\begin{figure*}
\begin{raggedright}
\bsent
   \litem{tags}\begin{tabular}[t]{ll}
      \tag{fc} & output of \cat{>B}$n$, some $n \geq 1$ (a forward composition rule) \\
      \tag{bc} & output of \cat{<B}$n$, some $n \geq 1$ (a backward composition rule) \\
      \tag{ot} & output of \cat{>B0} or \cat{<B0} (an application rule), or lexical item \\
   \end{tabular} 
   \litem{constrainedtemplate}\subsent
      \litem{constrainedtemplate:fa} Forward application \cat{>B0}: 
          {\alternatives{l}{x/y\tag{bc} \\
              x/y\tag{ot} \\}}\ 
\alternatives{l}{y\tag{fc} \\ 
              y\tag{bc} \\
              y\tag{ot} \\}
\ $\rightarrow$\ 
$x\tag{ot}$
      \item Backward application \cat{<B0}:
\alternatives{l}{y\tag{fc} \\ 
              y\tag{bc} \\
              y\tag{ot} \\}\ 
          {\alternatives{l}{x\bs y\tag{fc} \\ 
              x\bs y\tag{ot} \\}}
\ $\rightarrow$\ 
$x\tag{ot}$
\bigskip
      \item Fwd. composition \cat{>B}$n$ $(n \geq 1)$: 
          {\alternatives{l}{x/y\tag{bc} \\
              x/y\tag{ot} \\}}\ 
\alternatives{l}{y\slsh{n}z_n\cdots\slsh{2}z_2\slsh{1}z_1\tag{fc} \\ 
              y\slsh{n}z_n\cdots\slsh{2}z_2\slsh{1}z_1\tag{bc} \\
              y\slsh{n}z_n\cdots\slsh{2}z_2\slsh{1}z_1\tag{ot} \\}
$\rightarrow \begin{array}[t]{r}x\slsh{n}z_n\cdots\slsh{2}z_2\slsh{1}z_1\tag{fc} \\ \end{array}$
      \item Bwd. composition \cat{<B}$n$ $(n \geq 1)$:
\alternatives{l}{y\slsh{n}z_n\cdots\slsh{2}z_2\slsh{1}z_1\tag{fc} \\ 
              y\slsh{n}z_n\cdots\slsh{2}z_2\slsh{1}z_1\tag{bc} \\
              y\slsh{n}z_n\cdots\slsh{2}z_2\slsh{1}z_1\tag{ot} \\}\ 
          {\alternatives{l}{x\bs y\tag{fc} \\
              x\bs y\tag{ot} \\}}
$\rightarrow \begin{array}[t]{r}x\slsh{n}z_n\cdots\slsh{2}z_2\slsh{1}z_1\tag{bc} \\ \end{array}$
   \esubsent
\bigskip
   \litem{semtemplate}\subsent
      \litem{semtemplate:forward} Syn/sem for \cat{>B}$n$ $(n \geq 0)$: 
$\begin{array}[t]{c}
        x/y \\
        \Updownarrow \\
        f
\end{array}\ \ \ 
\begin{array}[t]{c}
       y\slsh{n}z_n\cdots\slsh{2}z_2\slsh{1}z_1 \\
       \Updownarrow \\
       g
\end{array}
\ \rightarrow\ 
\begin{array}[t]{c}
       x\slsh{n}z_n\cdots\slsh{2}z_2\slsh{1}z_1 \\
       \Updownarrow \\
       \lambda c_1 \lambda c_2 \ldots \lambda c_n . f(g(c_1)(c_2)\cdots(c_n))
\end{array}$

\bigskip

      \litem{semtemplate:backward} Syn/sem for \cat{<B}$n$ $(n \geq 0)$:
\begin{array}[t]{c}
       y\slsh{n}z_n\cdots\slsh{2}z_2\slsh{1}z_1 \\
       \Updownarrow \\
       g
\end{array}\ \ \ 
$\begin{array}[t]{c}
        x\bs y \\
        \Updownarrow \\
        f
\end{array}
\ \rightarrow\ 
\begin{array}[t]{c}
       x\slsh{n}z_n\cdots\slsh{2}z_2\slsh{1}z_1 \\
       \Updownarrow \\
       \lambda c_1 \lambda c_2 \ldots \lambda c_n . f(g(c_1)(c_2)\cdots(c_n))
\end{array}$
   \esubsent

\bigskip

   \litem{recipes}
a. \treeexample{1}\Trees{[ [ \cat{A/B} \cat{B/C/D} ].\cat{A/C/D} [ \cat{D/E} \cat{E/F} ].\cat{D/F} ].\cat{A/C/F}}
\hfill b. \treeexample{1}\Trees{[ [ [ \cat{A/B} \cat{B/C/D} ].\cat{A/C/D} \cat{D/E} ].\cat{A/C/E} \cat{E/F} ].\cat{A/C/F}}
\hfill c. \raisebox{-2\baselineskip}{\qroof{$\begin{array}{cccc}\cat{A/B} &\cat{B/C/D} &\cat{D/E} &\cat{E/F} \\
                                               {\scriptstyle f}     & {\scriptstyle g}      &  {\scriptstyle h}      &    {\scriptstyle k}
                              \end{array}$}.{$\begin{array}{c}
                                                 {\scriptstyle \lambda x \lambda y .f(g(h(k(x)))(y))} \\
                                                  \cat{A/C/F}
                                             \end{array}$} }

\esent
\end{raggedright}
\vspace{-1ex}   % tiny shortening of figure
\end{figure*}

It is trivial to modify any sort of CCG parser to find only the
normal-form parses.  No semantics is necessary; simply block any rule
use that would violate \refp{constraints}.  In general, detecting
violations will not hurt performance by more than a constant factor.  
Indeed, one might implement \refp{constraints} by modifying CCG's
phrase-structure grammar.  Each ordinary CCG category is
split into three categories that bear the respective tags from
\refp{tags}.  The 24 templates schematized in
\refp{constrainedtemplate} replace the two templates of
\refp{template}.  Any CFG-style method can still parse the resulting
spuriosity-free grammar, with tagged parses as in \refp{nfexamples}.
In particular, the polynomial-time, polynomial-space CCG chart parser
of \cite{VijayWeirPolyparseCorrect} can be trivially adapted to
respect the constraints by tagging chart entries.

It is interesting to note a rough resemblance between the tagged
version of CCG in \refp{constrainedtemplate} and the tagged Lambek
calculus {\bf L*}, which \cite{Hendriks93} developed to eliminate
spurious ambiguity from the Lambek calculus {\bf L}.  Although 
differences between CCG and {\bf L} mean that the details are quite
different, each system works by marking the output of certain rules, to
prevent such output from serving as input to certain other rules.

\subsection{Semantic equivalence}\label{semequiv}

We wish to establish that each semantic equivalence class contains
exactly one NF parse.  But what does ``semantically equivalent''
mean?  Let us adopt a standard model-theoretic view.

For each leaf (i.e., lexeme) of a given syntax tree, the lexicon
specifies a {\em lexical interpretation} from the model.  CCG then
provides a {\em derived interpretation} in the model for the complete
tree.  The standard CCG theory builds the semantics compositionally,
guided by the syntax, according to \refp{semtemplate}.  We may therefore
regard a syntax tree as a static ``recipe'' for combining word
meanings into a phrase meaning.

One might choose to say that two parses are semantically equivalent
iff they derive the same phrase meaning.  However, such a definition
would make spurious ambiguity sensitive to the fine-grained semantics
of the lexicon.  Are the two analyses of \cat{VP/VP $\,$VP $\,$VP\bs
VP} semantically equivalent?  If the lexemes involved are ``softly
knock twice,'' then yes, as softly(twice(knock)) and
twice(softly(knock)) arguably denote a common function in the semantic
model.  Yet for ``intentionally knock twice'' this is not the case: these
adverbs do not commute, and the semantics are distinct.

It would be difficult to make such subtle distinctions rapidly.  Let
us instead use a narrower, ``intensional'' definition of spurious
ambiguity.  The trees in (\ref{recipes}a--b) will be considered
equivalent because they specify the same ``recipe,'' shown in
(\ref{recipes}c).  No matter what lexical interpretations $f, g, h, k$
are fed into the leaves \cat{A/B}, \cat{B/C/D}, \cat{D/E}, \cat{E/F},
both the trees end up with the same derived interpretation, namely a
model element that can be determined from $f, g, h, k$ by calculating
$\lambda x \lambda y . f(g(h(k(x)))(y))$.

By contrast, the two readings of ``softly knock twice'' are considered
to be distinct, since the parses specify different recipes.  That is,
given a suitably free choice of meanings for the words, the two parses
can be made to pick out two different \cat{VP}-type functions in the
model.  The parser is therefore conservative and keeps both
parses.\footnote{\cite{Hepple+Morrill89,Hepple90thesis,Hendriks93}
appear to share this view of semantic equivalence.  Unlike
\cite{Karttunen86}, they try to eliminate only parses whose
denotations (or at least $\lambda$-terms) are systematically
equivalent, not parses that happen to have the same denotation through
an accident of the lexicon.}

\subsection{Normal-form parsing is safe \& complete}\label{sec:proofs}
The motivation for producing only NF parses (as defined by
\refp{constraints}) lies in the following existence and uniqueness
theorems for CCG.  

\begin{theorem}\label{thm:existence}  Assuming ``pure CCG,''
where all possible rules are in the grammar, any parse tree $\alpha$ is 
semantically equivalent to some NF parse tree ${\it NF}(\alpha)$. 
\end{theorem}
(This says the NF parser is {\em safe} for pure CCG: we will not lose
any readings by generating just normal forms.)

\begin{theorem}\label{thm:uniqueness}  Given distinct NF trees 
$\alpha \neq \alpha'$ (on the same sequence of leaves).  Then $\alpha$
and $\alpha'$ are not semantically equivalent. \end{theorem} (This
says that the NF parser is {\em complete}: generating only normal
forms eliminates {\em all} spurious ambiguity.)

Detailed proofs of these theorems are available on the cmp-lg archive,
but can only be sketched here.  Theorem~\ref{thm:existence} is proved
by a constructive induction on the order of $\alpha$, given below and
illustrated in \refp{pic:existence}:
\begin{itemize}
\item For $\alpha$ a leaf, put ${\em NF}(\alpha) = \alpha$.  

\item {\small ($\ang{R, \beta, \gamma}$ denotes the parse tree formed by
combining subtrees $\beta, \gamma$ via rule $R$.)} 

If $\alpha = \ang{R, \beta, \gamma}$, then take ${\em NF}(\alpha) =
\ang{R, {\em NF}(\beta), {\em NF}(\gamma)}$, which exists by inductive
hypothesis, unless this is not an NF tree.  In the latter case, WLOG, $R$ is a
forward rule and ${\em NF}(\beta) = \ang{Q, \beta_1, \beta_2}$ for
some forward composition rule $Q$.  Pure CCG turns out to provide
forward rules $S$ and $T$ such that $\alpha' = \ang{S, \beta_1, {\em
NF}(\ang{T,\beta_2,\gamma})}$ is a constituent and is semantically
equivalent to $\alpha$.  Moreover, since $\beta_1$ serves as the
primary subtree of the NF tree ${\em NF}(\beta)$, $\beta_1$ cannot be the
output of forward composition, and is NF besides.  Therefore
$\alpha'$ is NF: take ${\em NF}(\alpha) = \alpha'$.
\end{itemize}

\begin{raggedright}
\qtreecentertrue
\newsavebox{\subtree}
\savebox{\subtree}{\ontop{\Trees{[ $\beta_2$ $\gamma$ ].{$\stackrel{T}{\rightarrow}$}}}}  % need savebox because we can't call \Tree from inside itself (screws up stack)
\bsent
   \litem{pic:existence} 
      If ${\em NF}(\beta)$ not output of fwd. composition, \\ 
      $\alpha =$ 
      \makebox[1cm][c]{\ontop{\Trees{[ $\beta$ $\gamma$ ].{$\stackrel{R}{\rightarrow}$}}}}
      $\Longrightarrow$  
      \makebox[2cm][c]{\ontop{\Trees{[ ${\em NF}(\beta)$ ${\em NF}(\gamma)$ ].{$\stackrel{R}{\rightarrow}$}}}}
      $\stackrel{\mbox{def}}{=} {\em NF}(\alpha)$ \hspace{2em} 
      \\ \hspace*{\fill} 
   %%%%%%
      \hspace{-\leftmargin} else \hfill
      $\alpha =$ 
      \ontop{\Trees{[ $\beta$ $\gamma$ ].{$\stackrel{R}{\rightarrow}$}}}
      $\Longrightarrow$  
      \ontop{\Trees{[ ${\em NF}(\beta)$ $\gamma$ ].{$\stackrel{R}{\rightarrow}$}}}
      \\ \hspace{-\leftmargin}\raisebox{1ex}{$ =$}
      \makebox[1.75cm][c]{\ontop{\Trees{[ [ $\beta_1$ $\beta_2$ ].{$\stackrel{Q}{\rightarrow}$} $\gamma$ ].{$\stackrel{R}{\rightarrow}$}}}}
      \hfill \raisebox{1ex}{$\Longrightarrow$} \hfill
      \makebox[2.75cm][c]{\ontop{\Trees{[ $\beta_1$ {\({\em NF}\left(\makebox[1.2cm][c]{\usebox{\subtree}}\right)\)} ].{$\stackrel{S}{\rightarrow}$}}}}
      \hfill \raisebox{1ex}{$\stackrel{\mbox{def}}{=}$} \hfill {\em NF}$(\alpha)$ 
\esent
\qtreecenterfalse
\end{raggedright}
This construction resembles a well-known normal-form reduction
procedure that \cite{Hepple+Morrill89} propose (without proving
completeness) for a small fragment of CCG.

The proof of theorem~\ref{thm:uniqueness} (completeness) is longer and
more subtle.  First it shows, by a simple induction, that since $\alpha$
and $\alpha'$ disagree they must disagree in at least one of these ways:

\begin{itemize}
\item[(a)] There are trees $\beta, \gamma$ and rules $R \neq R'$ such
that $\ang{R, \beta, \gamma}$ is a subtree of $\alpha$ and $\ang{R',
\beta, \gamma}$ is a subtree of $\alpha'$.  (For example, \cat{S/S
S\bs S} may form a constituent by either \cat{<B1x} or \cat{>B1x}.)

\item[(b)] There is a tree $\gamma$ that appears as a subtree of 
both $\alpha$ and $\alpha'$, but combines to the left in one case 
and to the right in the other.
\end{itemize}
Either condition, the proof shows, leads to different ``immediate
scope'' relations in the full trees $\alpha$ and $\alpha'$ (in the
sense in which $f$ takes immediate scope over $g$ in $f(g(x))$ but not
in $f(h(g(x)))$ or $g(f(x))$). Condition (a) is straightforward.
Condition (b) splits into a case where $\gamma$ serves as a secondary
argument inside both $\alpha$ and $\alpha'$, and a case where it is a
primary argument in $\alpha$ or $\alpha'$.  The latter case requires
consideration of $\gamma$'s ancestors; the NF properties crucially
rule out counterexamples here.

The notion of scope is relevant because semantic interpretations for
CCG constituents can be written as restricted lambda terms, in such a
way that constituents having distinct terms must have different 
interpretations in the model (for suitable interpretations of the
words, as in \S\ref{semequiv}).  Theorem~\ref{thm:uniqueness} is
proved by showing that the terms for $\alpha$ and $\alpha'$ differ
somewhere, so correspond to different semantic recipes.

Similar theorems for the Lambek calculus were previously shown by
\cite{Hepple90thesis,Hendriks93}.  The present proofs for CCG
establish a result that has long been suspected: the spurious
ambiguity problem is not actually very widespread in CCG.
Theorem~\ref{thm:uniqueness} says {\em all} cases of spurious
ambiguity can be eliminated through the construction given in
theorem~\ref{thm:existence}.  But that construction merely ensures a
right-branching structure for ``forward constituent chains'' (such as
\cat{A/B $\,$B/C $\,$C} or \cat{A/B/C $\,$C/D $\,$D/E/F/G $\,$G/H}), and a left-branching
structure for backward constituent chains.  So these familiar chains
are the {\em only} source of spurious ambiguity in CCG.

\section{Extending the Approach to ``Restricted'' CCG}

The ``pure'' CCG of \S\ref{sec:nfparser} is a fiction.  Real CCG
grammars can and do choose a subset of the possible rules.  For
instance, to rule out \refp{badrule}, the (crossing) backward rule
\cat{N/N} \cat{N\bs N} $\rightarrow$ \cat{N/N} must be omitted from
English grammar.

\begin{raggedright}
\bsent
   \litem{badrule} [the\subcat{NP/N} [[big\subcat{N/N} [that likes John]\subcat{N\bs N}$\,$]\subcat{N/N} galoot\subcat{N}$\,$]\subcat{N}$\,$]\subcat{NP}
\esent
\end{raggedright}

If some rules are removed from a ``pure'' CCG grammar, some
parses will become unavailable.  Theorem~\ref{thm:uniqueness} remains
true ($\leq 1$ NF per reading).  Whether theorem~\ref{thm:existence}
($\geq 1$ NF per reading) remains true depends on what set of rules is
removed.  For most linguistically reasonable choices, the proof of
theorem~\ref{thm:existence} {\em will} go through,%
\footnote{For the proof to work, the rules $S$ and $T$ must be
available in the restricted grammar, given that $R$ and $Q$ are.  This
is usually true: since \refp{constraints} favors standard constituents
and prefers application to composition, most grammars will not block
the NF derivation while allowing a non-NF one.  (On the other hand,
the NF parse of \cat{A/B $\,$B/C $\,$C/D/E} uses \cat{>B2} twice, while the
non-NF parse gets by with \cat{>B2} and \cat{>B1}.)}
so that the normal-form parser of \S\ref{sec:nfparser} remains safe.
But imagine removing only the rule \cat{B/C $\,$C} $\rightarrow$
\cat{B}: this leaves the string \cat{A/B $\,$B/C $\,$C} with a
left-branching parse that has no (legal) NF equivalent.

In the sort of restricted grammar where theorem~\ref{thm:existence}
does {\em not} obtain, can we still find one (possibly non-NF) parse per equivalence
class?  Yes: a different kind of
efficient parser can be built for this case.

Since the new parser must be able to generate a non-NF parse when no
equivalent NF parse is available, its method of controlling spurious
ambiguity cannot be to enforce the constraints \refp{constraints}.
The old parser refused to build non-NF constituents; the new parser
will refuse to build constituents that are semantically equivalent to
already-built constituents.

This idea originates with \cite{Karttunen86}.  However, we can take
advantage of the core result of this paper, 
theorems~\ref{thm:existence} and~\ref{thm:uniqueness}, to do
Karttunen's redundancy check in $O(1)$ time---no worse than the
normal-form parser's check for \tag{fc} and \tag{bc} tags.  (Karttunen's
version takes worst-case exponential time for each redundancy check:
see footnote \S\ref{overview}.)

The insight is that theorems~\ref{thm:existence}
and~\ref{thm:uniqueness} establish a one-to-one map between semantic
equivalence classes and normal forms of the pure (unrestricted) CCG:

\begin{raggedright}
\bsent
   \litem{thm:onetoone} Two parses $\alpha, \alpha'$ of the pure CCG
are semantically equivalent iff they have the {\em same} normal form:
${\em NF}(\alpha) = {\em NF}(\alpha')$.
\esent
\end{raggedright}
The {\em NF} function is defined recursively by \S\ref{sec:proofs}'s
proof of theorem~\ref{thm:existence}; semantic equivalence is also
defined independently of the grammar.  So \refp{thm:onetoone} is meaningful and true
even if $\alpha, \alpha'$ are produced by a restricted CCG.  The tree
${\em NF}(\alpha)$ may not be a legal {\em parse} under the restricted
grammar.  However, it is still a perfectly good data structure that
can be maintained outside the parse chart, to serve as a magnet for
$\alpha$'s semantic class.  The proof of theorem~\ref{thm:existence}
(see \refp{pic:existence}) actually shows how to construct ${\em
NF}(\alpha)$ in $O(1)$ time from the values of ${\em NF}$ on smaller
constituents.  Hence, an appropriate parser can compute and cache the
NF of each parse in $O(1)$ time as it is added to the chart.  It can
detect redundant parses by noting (via an $O(1)$ array lookup) that
their NFs have been previously computed.

Figure~\refp{fig:algorithm} gives an efficient CKY-style algorithm
based on this insight.  (Parsing strategies besides CKY would also
work, in particular \cite{VijayWeirPolyparseCorrect}.)  The management of
cached NFs in steps~\ref{step:getnf}, \ref{step:savenf}, and
especially~\ref{step:usenf} ensures that duplicate NFs never enter the
{\em oldNFs} array: thus any alternative copy of $\alpha$.{\em nf} has
the same array coordinates used for $\alpha$.{\em nf} itself, because
it was built from identical subtrees.

The function {\tt PreferableTo}($\sigma$, $\tau$)
(step~\ref{step:pref}) provides flexibility about {\em which} parse
represents its class.  {\tt PreferableTo} may be defined at whim to
choose the parse discovered first, the more left-branching parse, or
the parse with fewer non-standard constituents.  Alternatively, {\tt
PreferableTo} may call an intonation or discourse module to pick the
parse that better reflects the topic-focus division of the sentence.
(A variant algorithm ignores {\tt PreferableTo} and constructs one
parse forest per reading.  Each forest can later be unpacked into
individual equivalent parse trees, if desired.)

\begin{figure*}
\begin{enumerate}
\setlength{\parskip}{-0.5ex}  
\item {\bf for} $i$ := 1 to {\tt n}
\item \tb  $C[i-1,i]$ := ${\tt LexCats}({\tt word}[i])$ \hfill (* word $i$ stretches from point $i-1$ to point $i$ *)
\item {\bf for} $width$ := 2 to {\tt n}
\item \tb  {\bf for} $start$ := 0 to ${\tt n}-width$
\item \tb  \tb  $end$ := $start+width$
\item \tb  \tb  {\bf for} $mid$ := $start+1$ to $end-1$
\item \tb  \tb  \tb {\bf for} each parse tree $\alpha = \ang{R, \beta, \gamma}$ that could be formed by combining some 
\item[]\tb \tb  \tb $\beta \in C[start, mid]$ with some $\gamma \in C[mid, end]$ by a rule $R$ of the (restricted) grammar
\litem{step:makenf}
      \tb  \tb  \tb  \tb  $\alpha.{\em nf}$ := ${\tt NF}(\alpha)$ \hfill (* can be computed in constant time using the $.{\em nf}$ fields of $\beta$, $\gamma$, and
\item[] \hfill \tb  other constituents already in $C$.  Subtrees are also NF trees. *)
\litem{step:getnf}
      \tb  \tb  \tb  \tb  ${\em existingNF}$ := ${\em oldNFs}[\alpha.{\em nf.rule}, \alpha.{\em nf.leftchild.seqno}, \alpha.{\em nf.rightchild.seqno}]$
\item \tb  \tb  \tb  \tb  {\bf if} {\bf undefined}({\em existingNF}) \hfill (* the first parse with this NF *)
\item \tb  \tb  \tb  \tb  \tb  $\alpha$.{\em nf.seqno} := ({\em counter} := ${\em counter} + 1$) \hfill (* number the new NF \& add it to {\em oldNFs} *)
\litem{step:savenf}
      \tb  \tb  \tb  \tb  \tb  ${\em oldNFs}[\alpha.{\em nf}.rule, \alpha.{\em nf.leftchild.seqno}, \alpha.{\em nf.rightchild.seqno}]$ := $\alpha$.{\em nf}
\item \tb  \tb  \tb  \tb  \tb  add $\alpha$ to $C[start,end]$
\item \tb  \tb  \tb  \tb  \tb  $\alpha$.{\em nf.currparse} := $\alpha$
\litem{step:pref}
      \tb  \tb  \tb  \tb  {\bf elsif} ${\tt PreferableTo}(\alpha, {\em existingNF}.currparse)$  \hfill (* replace reigning parse? *)
\litem{step:usenf} 
      \tb  \tb  \tb  \tb  \tb  $\alpha$.{\em nf} := ${\em existingNF}$  \hfill (* use cached copy of NF, not new one *)
\item \tb  \tb  \tb  \tb  \tb  remove $\alpha$.{\em nf.currparse} from $C[start,end]$
\item \tb  \tb  \tb  \tb  \tb  add $\alpha$ to $C[start,end]$
\item \tb  \tb  \tb  \tb  \tb  $\alpha$.{\em nf.currparse} := $\alpha$
\item {\bf return}(all parses from $C[0,n]$ having root category \cat{S})
\end{enumerate}
\caption{Canonicalizing CCG parser that handles arbitrary restrictions on the rule set.  (In practice, a simpler normal-form parser will suffice for most grammars.)}\label{fig:algorithm}
\end{figure*}

\cite{VijayWeirPolyparseSpur} also give a method for removing ``one
well-known source'' of spurious ambiguity from restricted CCGs;
\S\ref{sec:proofs} above shows that this is in fact the only source.
However, their method relies on the grammaticality of certain
intermediate forms, and so can fail if the CCG rules can be {\em
arbitrarily} restricted.  In addition, their method is less efficient
than the present one: it considers parses in pairs, not singly, and
does not remove any parse until the entire parse forest has been built.

\section{Extensions to the CCG Formalism}

In addition to the {\bf B}$n$ (``generalized composition'') rules
given in \S\ref{ccgdefs}, which give CCG power equivalent to TAG,
rules based on the {\bf S} (``substitution'') and {\bf T}
(``type-raising'') combinators can be linguistically useful.  {\bf S}
provides another rule template, used in the analysis of parasitic gaps
\cite{SteedmanParasitic,Szabolcsi}:

\begin{raggedright}
\bsent
  \litem{subst}\subsent
     \litem{subst:forward} \cat{>S}:
\begin{array}[t]{c}
       x/y\slsh{1}z \\
       \Updownarrow \\
       f
\end{array}\ \ \ 
$\begin{array}[t]{c}
        y\slsh{1}z  \\
        \Updownarrow \\
        g
\end{array}
\ \rightarrow\ 
\begin{array}[t]{c}
       x\slsh{1}z \\
       \Updownarrow \\
       \lambda z . f(z)(g(z)) 
\end{array}$
     \litem{subst:backward} \cat{<S}: \hfill$y\slsh{1}z\hspace{0.8em}x\bs y\slsh{1}z\hspace{0.8em}\rightarrow\hspace{4mm}x\slsh{1}z$\hspace{1em}\hfill\mbox{}
   \esubsent
\esent
\end{raggedright}

Although {\bf S} interacts with {\bf B}$n$ to produce another source
of spurious ambiguity, illustrated in \refp{Sambig}, the additional
ambiguity is not hard to remove.  It can be shown that when the
restriction \refp{Sconstraints} is used together with
\refp{constraints}, the system again finds exactly one parse from
every equivalence class.

\begin{raggedright}
\bsent
   \litem{Sambig}
\smallskip
      \subsent
      \item\mbox{}\vspace{-\baselineskip} \\ \mbox{}\hspace{-2\leftmargin}%
\Trees{[ [ \catword{VP$_2$/NP}{filed} \catword{VP$_1$\bs VP$_2$/NP}{[without-reading]} ].\catrule{VP$_1$/NP}{<Sx} \catword{VP$_0$\bs VP$_1$}{yesterday} ].\catrule{VP$_0$/NP}{<Bx}} \\
\bigskip
      \litem{Sambig:nonnf}\mbox{}\vspace{-\baselineskip} \mbox{}\hspace{-2\leftmargin}% \\ 
\Trees{[ \cat{VP$_2$/NP} [ \cat{VP$_1$\bs VP$_2$/NP}  \cat{VP$_0$\bs VP$_1$} ].\catrule{VP$_0$\bs VP$_2$/NP}{<B2} ].\catrule{VP$_0$/NP}{<Sx}} \\
   \esubsent
\bigskip
   \litem{Sconstraints}\subsent
       \litem{Sconstraints:forward} No constituent produced by \cat{>B}$n$, any $n \geq 2$, ever serves
as the primary (left) argument to \cat{>S}.
       \litem{Sconstraints:backward} No constituent produced by \cat{<B}$n$, any $n \geq 2$, ever serves as the primary (right) argument to \cat{<S}.

   \esubsent
\esent
\end{raggedright}

Type-raising presents a greater problem.  Various new spurious
ambiguities arise if it is permitted freely in the grammar.  In
principle one could proceed without grammatical type-raising:
\cite{DowtyLexTR,SteedmanLexTR} have argued on linguistic grounds that
type-raising should be treated as a mere lexical redundancy property.
That is, whenever the lexicon contains an entry of a certain category
\cat{X}, with semantics $x$, it also contains one with (say) category
\cat{T/(T\bs X)} and interpretation $\lambda p.p(x)$.  As one might
expect, this move only sweeps the problem under the rug.  If type-raising
is lexical, then the definitions of this paper do not recognize \refp{trbad1}
as a spurious ambiguity, because the two parses are now, technically speaking,
analyses of different sentences.  Nor do they recognize the redundancy
in \refp{trbad2}, because---just as for the example ``softly knock
twice'' in \S\ref{semequiv}---it is contingent on a kind of lexical
coincidence, namely that a type-raised subject commutes with a
(generically) type-raised object.  Such ambiguities are left to future
work.

\begin{raggedright}
\bsent
   \litem{trbad1} {\small [John\subcat{NP} left\subcat{S\bs NP}]\subcat{S} vs. [John\subcat{S/(S\bs NP)} left\subcat{S\bs NP}]\subcat{S}}
   \litem{trbad2} {\small [S/(S\bs {\tiny NP}\subcat{S}) [S\bs {\tiny NP}\subcat{S}/{\tiny NP}\subcat{O}/{\tiny NP}\subcat{I} {\bf T}\bs ({\bf T}/{\tiny NP}\subcat{O})]]\subcat{S/S\subcat{I}} \\ \makebox[0in][r]{vs. }[S/(S\bs {\tiny NP}\subcat{S}) S\bs {\tiny NP}\subcat{S}/{\tiny NP}\subcat{O}/{\tiny NP}\subcat{I}] {\bf T}\bs ({\bf T}/{\tiny NP}\subcat{O})]\subcat{S/S\subcat{I}}}
\esent
\end{raggedright}

\section{Conclusions}

The main contribution of this work has been formal: to establish a
normal form for parses of ``pure'' Combinatory Categorial Grammar.
Given a sentence, every reading that is available to the grammar has
exactly one normal-form parse, no matter how many parses it has {\em
in toto}.

A result worth remembering is that, although TAG-equivalent CCG allows
free interaction among forward, backward, and crossed composition
rules of any degree, two simple constraints serve to eliminate all
spurious ambiguity.  It turns out that all spurious ambiguity arises from 
associative ``chains'' such as \cat{A/B $\,$B/C $\,$C} or \cat{A/B/C $\,$C/D
$\,$D/E\bs F/G $\,$G/H}.  \cite{Wittenburg87,Hepple+Morrill89} anticipate this
result, at least for some fragments of CCG, but leave the proof to
future work.

These normal-form results for pure CCG lead directly to useful parsers
for real, restricted CCG grammars.  Two parsing algorithms
have been presented for practical use.  One algorithm finds only
normal forms; this simply and safely eliminates spurious ambiguity
under {\em most} real CCG grammars.  The other, more complex algorithm
solves the spurious ambiguity problem for {\em any} CCG grammar, by
using normal forms as an efficient tool for grouping semantically
equivalent parses.  Both algorithms are safe, complete, and efficient.

In closing, it should be repeated that the results provided are for
the TAG-equivalent {\bf B}$n$ (generalized composition) formalism of
\cite{Joshi+al}, optionally extended with the {\bf S} (substitution)
rules of \cite{Szabolcsi}.  The technique eliminates all spurious
ambiguities resulting from the interaction of these rules.  Future
work should continue by eliminating the spurious ambiguities that
arise from grammatical or lexical type-raising.

\end{document}